\documentclass{elsarticle}

\usepackage{macros}
\usepackage{amsmath,amssymb}
\usepackage{multicol}
\usepackage{tikz}
\usepackage{tikz-cd}
\usepackage{verbatim}
\usepackage{wrapfig}
\usetikzlibrary{shapes, snakes, backgrounds, calc}
\usetikzlibrary{decorations.pathmorphing}
\usepackage{transparent}
\usepackage{graphicx}
\usepackage[outline]{contour}
\newcommand{\sem}[1]{[\![{#1}]\!]}
\newcommand{\cev}[1]{\reflectbox{\ensuremath{\vec{\reflectbox{\ensuremath{#1}}}}}}

\makeatletter
\tikzset{circle split part fill/.style  args={#1,#2}{%
 alias=tmp@name, 
  postaction={%
    insert path={
     \pgfextra{%
     \pgfpointdiff{\pgfpointanchor{\pgf@node@name}{center}}%
                  {\pgfpointanchor{\pgf@node@name}{east}}%
     \pgfmathsetmacro\insiderad{\pgf@x}
      \fill[#1] (\pgf@node@name.base) ([xshift=-\pgflinewidth]\pgf@node@name.east) arc
                          (0:180:\insiderad-\pgflinewidth)--cycle;
      \fill[#2] (\pgf@node@name.base) ([xshift=\pgflinewidth]\pgf@node@name.west)  arc
                           (180:360:\insiderad-\pgflinewidth)--cycle;            
         }}}}}  
 \makeatother

\begin{document}

\title{Knowledge representation and update in hierarchies of graphs}
\author{Russ Harmer}\ead{russell.harmer@ens-lyon.fr}
\author{Eugenia Oshurko}\ead{ievgeniia.oshurko@ens-lyon.fr}
\address{Univ Lyon, EnsL, UCBL, CNRS, LIP, F-69342 LYON Cedex 07, France}

\begin{abstract}
A mathematical theory is presented for the representation of knowledge in the form of a directed acyclic hierarchy of objects in a category where all paths between any given pair of objects are required to be equal. The conditions under which knowledge update, in the form of the sesqui-pushout rewriting of an object in a hierarchy, can be propagated to the rest of the hierarchy, in order to maintain all required path equalities, are analysed: some rewrites must be propagated forwards, in the direction of the arrows, while others must be propagated backwards, against the direction of the arrows, and, depending on the precise form of the hierarchy, certain composability conditions may also be necessary. The implementation of this theory, in the \texttt{ReGraph} Python library for (simple) directed graphs with attributes on nodes and edges, is then discussed in the context of two significant use cases.
\end{abstract}


\maketitle

\section{Introduction}

We present a generic framework for knowledge representation (KR) based on hierarchies of objects from an appropriately structured category: a hierarchy is a directed acyclic graph (DAG) whose nodes are \emph{objects} of the category and whose edges are \emph{arrows} of the category 
such that all paths between each pair of objects are equal; we refer to this as the \emph{commutativity} condition.

The principal model of interest to us in this paper uses (simple) graphs and homomorphisms so that a hierarchy is a DAG whose nodes are themselves (simple) graphs. In this model, an edge of the DAG $h : G \rightarrow T$ asserts that the graph $G$ is \emph{typed} by $T$, \ie $T$ defines the kinds of nodes and kinds of edges (and attributes, if desired) that exist in $G$ and $h$ specifies, for each node and edge (and attribute) of $G$, which kind it is. As such, $T$ can be viewed as a more abstract representation of knowledge of which $G$ provides a more concrete instantiation.

We require certain structure on the category in order to be able to perform sesqui-pushout rewriting \cite{corradini2006sesqui} to update an object in the hierarchy. However, such an update may invalidate some of the typing arrows of the hierarchy. The main contribution of this paper is to present a mathematical theory that guarantees the reconstruction of a valid hierarchy, after an arbitrary rewrite of an object, by appropriately \emph{propagating} that rewrite to the other objects in the hierarchy. In general, this only concerns a subgraph of the hierarchy that is determined as a function of the nature of the update and the paths to and/or from the updated object. In the case where there are multiple paths between a given pair of objects of the hierarchy, this reconstruction depends on the satisfaction of a \emph{composability} condition, guaranteeing that the propagated rewrites are compatible, in order to maintain validity of the commutativity condition.

\subsection*{Graph databases}

Modern database systems are increasingly migrating towards graph-based representations as a response to the growing wealth of data---from domains as varied as social or transport networks, the semantic web or biological interaction networks---that are most naturally expressed in those terms. However, unlike traditional relational DBs or earlier graph-based formats such as RDF, most graph DBs based on the richer model of property graphs \cite{francis2018cypher,bonifati2019schema} do not provide a native notion of \emph{schema}. Our notion of hierarchy provides a mathematical framework for this. Indeed, an explicitly given schema graph to which a data, or instance, graph is homomorphic is the simplest non-trivial example of a hierarchy in our sense: the nodes of the schema specify the types of entites allowed in the system; its edges specify which edges between different types of nodes are allowed; and the attributes on its nodes and edges define the set of permitted attributes for nodes and edges. As such, the existence of a homomorphism from a data graph to a schema graph provides a proof of schema \emph{validation} \cite{bonifati2019schema}.

Our theory of propagation of rewriting in a hierarchy precisely captures the ways in which schema-aware DBs can be updated: a \emph{descriptive} update occurs when the data is modified and the schema has to adjust accordingly; while a \emph{prescriptive} update occurs when the schema is modified and the data needs to be adjusted \cite{bonifati2019schema}. More precisely, if we \emph{add} a node to the data graph and choose not to specify that its type already exists in the schema graph, in order to maintain the homomorphism from data to schema, we must propagate this operation to the schema graph to create a new node in the schema graph to type the new node of the data graph; similarly, if we \emph{merge} two nodes of different types of the data graph, we must merge the corresponding typing nodes of the schema. Conversely, if we \emph{delete} a node of the schema graph, we can only maintain the homomorphism by deleting all instances of that node in the data graph; and if we \emph{clone} a node of the schema and choose not specify how to retype its instances in the data graph, those instances must be cloned in the data graph. In summary, \emph{add} and \emph{merge} updates propagate \emph{forwards}, in the direction of the typing homomorphism, while \emph{clone} and \emph{delete} updates propagate \emph{backwards}; and, as we will show, these observations remain true for general hierarchies.

Our theory thus provides a specification of how to enforce an \emph{abstraction barrier} on a schema-less graph DB that provides the illusion of being schema-aware. Our Python library \texttt{ReGraph} implements this for the Neo4j graph DB by fixing an encoding of the data and schema graphs and the typing homomorphism within the single graph provided by Neo4j and translating any combination of clone, delete, add and merge operations into a corresponding query written in the Cypher query language used by Neo4j \cite{bonifati2019schema}. 
More importantly, our theory also provides a specification of how to enforce the abstraction barrier for an \emph{arbitrary} hierarchy---modulo the need to fix the encoding into Neo4j and the translation of update operations into Cypher. However, these two requirements are generic and can be derived systematically. As such, we provide the foundations for exploiting Neo4j (or similar graph DBs) as a platform for arbitrary, user-defined graph-based KR systems.

\subsection*{The KAMI bio-curation tool}

The core of the \texttt{KAMI} bio-curation system \cite{harmer2019bio} has a richer 3-level hierarchy. At the root lies its \emph{meta-model}, a fixed, hard-wired graph which defines the universe of discourse pertinent to the rule-based modelling of protein-protein interactions (PPIs) in cellular signalling: genes, regions of genes, binding and enzymatic actions, \etc The meta-model types an \emph{action graph} which defines the particular collection of genes (and so on) of interest to a \emph{corpus} of knowledge, \eg a signalling pathway. The action graph types a collection of \emph{nugget} graphs, each representing the detailed conditions needed for a particular PPI to occur. In other words, an action graph summarizes the \emph{anatomy} of a system while the collection of nugget graphs provides a representation of the \emph{physiology} that determines how the system can behave.

In general, an update of a nugget graph refers to some anatomic features that already exist in the action graph and to others that must be added to maintain typing; this is performed automatically by forward propagation. It is important that propagation does not continue to the meta-model (which must remain unchanged); this is achieved by requiring that all new anatomic features specify (at least) how they are to be typed by the meta-model. This gives an example of the notion of \emph{controlled} forward propagation, as discussed in section 3, and can be seen as a more general instance of a descriptive DB update which actually preserves the current schema.

A knowledge corpus in \texttt{KAMI} can be contextualized, with respect to a choice of gene products, through an update of its action graph, giving rise to what we call a \texttt{KAMI} \emph{model}; in the terminology of DBs, this is analogous to a \emph{materialized view}---a contextualized copy of part of the original DB that can be manipulated independently. The effect of this update propagates backwards to the nugget graphs. This propagation is not controlled---the cloning of a gene precisely gives rise to multiple gene products---unlike the case of \emph{concept refinement} where the cloning of a schema node is accompanied by a specification of how to retype all instances of the original node in the data graph in terms of the refined schema. We discuss backward propagation, including the controlled case, in section 4.

\subsection*{Related work}
Slice categories provide many rich models of \emph{typed} sesqui-pushout rewriting \cite{corradini2006sesqui}, \eg $\Set / T$ defines a setting for multi-set rewriting over the set $T$. We provide a powerful generalization of this where, through the use of a hierarchy, we can not only guarantee that rewriting an object always returns a well-typed result but, additionally, can dynamically modify the typing object $T$. Our approach is related to the change-of-base functor familiar from algebraic topology and to its right adjoint whose existence characterizes pullback complements \cite{dyckhoff1987exponentiable}. Indeed, in a sense, our work can be seen as providing a means of exploiting this theory, in a form that can be used for knowledge representation and graph databases, even when only those PBCs required for SqPO rewriting exist.

The arrows in our hierarchies correspond intuitively to the type, or instance-of, relationships found in entity-relationship (ER) modelling \cite{chen1976entity} or UML, \ie they are relations that cross from one meta-model layer to another. They also generally correspond to TBox statements in Description Logic \cite{baader2003description} although, in some cases, this intuition breaks down since an object, such as the nugget graph of \texttt{KAMI}, with no incoming arrows usually corresponds to a collection of ABox statements about instances of the concepts defined below it in the hierarchy. In this paper, we do not consider the specialization/generalization, or is-a, relationships found in ER modelling for the reason that the rewrite of an object does not need to propagate across such relations. 

\section{Preliminaries}

In this section, we discuss the necessary preliminary material concerning graph rewriting---specifically the definitions of pullback complements and image factorizations---and provide a formal definition of our notion of hierarchy.

Let us begin by defining a piece of useful terminology. We use the term \emph{element} to refer to any concrete constituent of an object in a concrete category of interest to us, \eg an element (in the usual sense) of a set or a node, edge or attribute of a graph.

\subsection{Sesqui-pushout rewriting}

Sesqui-pushout (SqPO) rewriting \cite{corradini2006sesqui} is a generalization of double pushout (DPO) and single pushout (SPO) rewriting \cite{corradini1997algebraic,ehrig1997algebraic}. In typical concrete settings, it allows for the expression of rules for all elementary manipulations generally considered in traditional graph (or multi-set) rewriting: the addition, deletion, merging and cloning of elements as well as the modification of the value(s) associated with an attribute. It extends SPO rewriting, by allowing for cloning, which in turn extends DPO rewriting by allowing for side-effects due to deletion (but not those due to merging, which DPO rewriting already accommodates).

The abstract formulation of SqPO rewriting requires the categorical notion of final pullback complements (PBCs) \cite{dyckhoff1987exponentiable}. As this remains (slightly) non-standard, we include a full definition here.

Given a pair of composable arrows $f : A \rightarrow B$ and $g : B \rightarrow D$, their \emph{final pullback complement} is a pair of composable arrows $\hat{g} : A \rightarrow C$ and $\hat{f} : C \rightarrow D$ such that the resulting square is a pullback (PB) satisfying the following universal property (UP): given a PB square (using $g$ but not necessarily $f$)
\[
	\begin{tikzcd}[ampersand replacement=\&]
	B \arrow[d, "g"'] \& \arrow[l, "f'"'] \arrow[d, "\hat{g}'"] A' \\
	D \& \arrow[l, "\hat{f}'"] C'
	\end{tikzcd}
\]
and an arrow $h : A' \rightarrow A$ such that $f' = f \circ h$, there exists a unique arrow $\hat{h} : C' \rightarrow C$ such that $\hat{f}' = \hat{f} \circ \hat{h}$ and $\hat{g} \circ h = \hat{h} \circ \hat{g}'$.
\[
	\begin{tikzcd}[ampersand replacement=\&]
	\& \& A' \arrow[dll, bend right=20, "f'"'] \arrow[d, "\hat{g}'", pos=0.4] \arrow[dl, "h" description] \\
	B \arrow[d, "g"'] \& \arrow[l, "f"'] \arrow[d, "\hat{g}"] A \& C' \arrow[dll, bend left=65, "\hat{f}'", pos=0.3] \arrow[dl, dotted, "\hat{h}" description] \\
	D \& \arrow[l, "\hat{f}"] C
	\end{tikzcd}
\]

SqPO rewriting can be performed in any category with all PBs, all PBCs over monos, \ie where $g : B \rightarrow D$ is a mono, and all pushouts (POs); we further require that POs preserve monos. These conditions are satisfied in all concrete settings of interest to us, typically sets and (simple) graphs with attributes, and potentially in many other concrete settings to which our theory would therefore also apply.

In order to perform SqPO rewriting of a single object, we only actually need the existence of PBs and POs of (co-)spans where one arrow is a mono. However, in this paper, we sometimes have need of more general PBs and POs to express the propagation of rewriting through a hierarchy. We also need the existence of all \emph{image factorizations} (IFs). As this notion is not standard in graph rewriting, we give an explicit definition of its UP.

The image factorization of an arrow $f : A \rightarrow B$ is a mono $m : I \rightarrowtail B$ such that (i) there exists an arrow $e : A \rightarrow I$ such that $f = m \circ e$; and (ii) for any arrow $e' : A \rightarrow I'$ and mono $m' : I' \rightarrowtail B$ such that $f = m' \circ e'$, there exists a unique arrow $i : I \rightarrow I'$ such that $m = m' \circ i$.
\[
	\begin{tikzcd}[ampersand replacement=\&]
		A \arrow[ddr, bend right, "e'"'] \arrow[dr, "e"'] \arrow[rr, "f"] \& \& B \\
		\& I \arrow[ur, tail, "m"'] \arrow[d, dashed, "i", pos=0.3] \\
		\& I' \arrow[uur, bend right, tail, "m'"']
	\end{tikzcd}
\]
In the concrete settings of interest to us, the IF of an arrow coincides with the familiar notion of its epi-mono factorization. However, we have no (abstract) need for the first arrow to be an epi and so prefer the more abstract requirement of having IFs of all arrows.

We consider a \emph{rule} to be simply an arrow. A \emph{restrictive instance} of a rule $r^- : L \leftarrow P$ in an object $G$ is a mono $m : L \rightarrowtail G$ from the target object $L$; in this case, we refer to $L$ as the LHS and $P$ as the RHS of $r^-$. An \emph{expansive instance} of a rule $r^+ : P \rightarrow R$ is a mono from the source object $P$; in this case, we refer to $P$ as the LHS and $R$ as the RHS of $r^+$.

The usual notion of rule, \ie a span of arrows, consists of two rules, $r^-$ and $r^+$, in our sense with a common source object $P$. Given a restrictive instance $m$ of the first, the PBC of $r^-$ and $m$ provides an expansive instance $m^-$ of the second and the PO of $m^-$ and $r^+$ completes the overall rewrite of $G$ to $G^+$.
\[
	\begin{tikzcd}[ampersand replacement=\&]
		L \arrow[d, tail, "m"'] \& P \arrow[l, "r^-"'] \arrow[d, tail, "m^-"]  \arrow[r, "r^+"] \& R \arrow[d, tail, "m^+"]  \\
		G \& G^- \arrow[l, "g^-"] \arrow[r, "g^+"'] \& G^+
	\end{tikzcd}
\]

\subsection{Hierarchies}

We formalize the notion of hierarchy by first defining the underlying skeleton of the KR as a DAG then defining the hierarchy itself as a graph homomorphic to the skeleton. More precisely, a \emph{skeleton} is a directed acyclic simple graph and a \emph{hierarchy} over the skeleton $\mathcal{S}$ is a directed simple graph $\mathcal{H}$ equipped with a homomorphism to $\mathcal{S}$; as such, it is also directed acyclic.

The simplest non-trivial skeleton consists of two nodes, $d$ and $s$, with a single edge $e : d \rightarrow s$ between them; this expresses that there are two types of object---let us call them \emph{data} and \emph{schema} nodes---and that data nodes have edges to schema nodes. The simplest hierarchy over this skeleton is the skeleton itself; this defines a KR system that contains a single data node, a single schema node and an edge from the former to the latter. However, a hierarchy over this skeleton could contain multiple data and/or schema nodes where a single data node may have edges to multiple schema nodes and/or multiple data nodes may have edges to a single schema node.

More generally, a skeleton specifies the \emph{types} of nodes and edges that can exist, \ie the \emph{shape} of the KR, while a hierarchy specifies the \emph{instances} of those objects and arrows that actually exist, \ie the current \emph{structure} of the KR. The skeleton of a KR system generally remains invariant throughout its lifetime while its hierarchy evolves over time. However, in this paper, we do not consider operations that modify the structure of a hierarchy; instead, we are interested in instantiating hierarchies with content and in modifying that content.

The \emph{instantiation} of a hierarchy $\mathcal{H}$ in a category $\C$ is defined by assigning an object $\sem{n}$ of $\C$ to each node $n$ of the hierarchy and an arrow $\sem{e} : \sem{n_1} \rightarrow \sem{n_2}$ of $\C$ to each edge $e : n_1 \rightarrow n_2$ of the hierarchy in such a way that the commutativity condition is satisfied. (This can be seen as a functor from the reflexive, transitive closure of $\mathcal{H}$---defined in such a way as to be still a simple graph---to $\C$.)

 An instantiation of the above hierarchy $e : d \rightarrow s$ in $\Set$ therefore consists of two sets, $G$ and $T$, and a function $h : G \rightarrow T$ between them; this can be seen as an intensional representation of a multi-set where $G$ defines the individuals, $T$ defines the types of individuals and $h$ assigns a type to each individual. We are interested in operations that update (one or other of) these two sets, \eg adding an element to $G$ or removing an element from $T$. However, such operations necessitate the making of changes to the function $h$ and, potentially, to the other set as well, \eg if we add an element to $G$, we must update $h$ to be defined on that new element---and this may entail adding a new element to $T$, if it does not already contain the desired type.
 

Our theory provides a general framework for expressing and applying such updates of objects, as SqPO rules, and determines how the arrows and other objects must be updated in consequence in order to maintain a valid instance of the hierarchy.
In the next two sections, we explain how (i) an expansive rewrite of $G$ is \emph{propagated} to $T$ in order to obtain a rewritten hierarchy $h^+ : G^+ \rightarrow T^+$; and (ii) a restrictive rewrite of $T$ is propagated to $G$ in order to obtain a rewritten $h^- : G^- \rightarrow T^-$.

\section{Forward propagation}

Throughout this section and the next, we consider two objects $G$ and $T$ and an arrow $h : G \rightarrow T$ of a category $\C$ possessing all the structure required for SqPO rewriting, \ie an instantiation in $\C$ of the hierarchy $e : d \rightarrow s$.

In this section, we consider a rule $r : L \rightarrow L^+$ and an expansive instance $m : L \rightarrowtail G$ of $r$ in $G$. Note that we immediately obtain a typing of $L$ by $T$ by composition, \ie $h \circ m : L \rightarrow T$.

\subsection{The strict phase of forward rewriting}

In order to decide how to propagate a rewrite of $G$ to $T$, we must further specify to what extent we wish to consider the RHS $L^+$ of $r$ to be typed by $T$. There are two extreme cases: the first is where we provide an arrow from $L^+$ to $T$, \ie $L^+$ is itself typed by $T$; the other is the case where nothing in the complement of the image of $r$ is homomorphic to $T$. In the first case, which we call a \emph{strict} rewrite of $G$, the rewritten $G^+$ is still typed by $T$; in the other case, which we call the \emph{canonical} propagation to $T$, we must propagate all changes in $G$ to $T$. In between these extremes, we can specify those elements, not in the image of $r$, that we nonetheless wish to be typed by $T$.

\paragraph{Definition}
Given a rule $r : L \rightarrow L^+$, a \emph{forward factorization} of $r$ is an object $L'$ and arrows $r' : L \rightarrow L'$ and $r^+ : L' \rightarrow L^+$ such that $r = r^+ \circ r'$; and an arrow $x : L' \rightarrow T$ such that $h \circ m = x \circ r'$.
\begin{equation}\label{f-fact}
\begin{tikzcd}[ampersand replacement=\&]
	L \arrow[d, "h \circ m"'] \arrow[r, "r"] \arrow[dr, red, "r'" description] \& L^+ \\ 
	T \& \color{red} L' \arrow[u, red, "r^+"'] \arrow[l, red, "x"] 
\end{tikzcd}
\end{equation}

In the case of strict rewriting, $L'$ is isomorphic to $L^+$ so that $x : L^+ \rightarrow T$ whereas, if $L'$ is isomorphic to $L$, $x$ specifies nothing more than $h \circ m$. In the concrete settings of multi-sets and of graphs, $r'$ is frequently taken to be a mono, \ie it expresses a rule that only \emph{adds} elements that can be typed by $T$, but in the abstract setting we have no need to enforce this as a requirement.

The factorization of $r$ splits its application into two phases: the \emph{strict} phase, specified by $r'$, which modifies only $G$; and the \emph{canonical} phase, specified by $r^+$, which modifies $G$ and $T$.

\paragraph{Definition}
The \emph{strict rewrite} of $G$ is defined by taking the PO of $m$ and $r'$. By the definition (\ref{f-fact}) of forward factorization and the universal property of this PO, we obtain a (unique) arrow $h'$ that types $G'$ by $T$. Note that $x = h' \circ m'$.
\begin{equation}\label{f-strict}
\begin{tikzcd}[ampersand replacement=\&]
	L \arrow[d, tail, "m"'] \arrow[r, "r'"] \& L' \arrow[d, blue, tail, "m'"] \arrow[dd, bend left=50, "x"] \\
	G \arrow[dr, "h"'] \arrow[r, blue, "g'"] \& \color{blue} G' \arrow[d, dashed, "h'"] \\
	\& T
\end{tikzcd}
\end{equation}

Note that the strict rewrite can only merge elements of $G$ that have the same type; this is a consequence of the requirement that $h \circ m = x \circ r'$. It can also add multiple elements to $G$---provided they can all be typed in $T$.

This strict phase of rewriting was discussed briefly in \cite{harmer2017hdr} as being the only kind of rewrite that can be performed if $T$ is hard-wired as the base object of a slice category; typically, a descriptive update that \emph{preserves} the current schema.

\subsection{The canonical phase of forward propagation}

Our more general and flexible setting of hierarchies enables a second phase of rewriting where the remaining changes to be made to $G'$, as specified by $r^+$, are additionally propagated to $T$, \ie the base object changes.

\paragraph{Definition}
The \emph{rewrite} of $G$ is completed by taking the PO of $r^+$ and $m'$. The \emph{forward propagation} to $T$ is then defined by taking the PO of $g^+$ and $h'$. The final typing of $G^+$ by $T^+$ is given by $h^+$.
\begin{equation}\label{f-canonical}
\begin{tikzcd}[ampersand replacement=\&]
	 L'\arrow[d, tail, "m'"'] \arrow[r, "r^+"] \& L^+ \arrow[d, blue, tail, "m^+"] \& \\
	 G' \arrow[r, blue, "g^+"'] \& \color{blue} G^+ \&
\end{tikzcd}
\begin{tikzcd}[ampersand replacement=\&]
	 G' \arrow[d, "h'"'] \arrow[r, "g^+"] \& G^+ \arrow[d, blue, "h^+"] \\
	 T \arrow[r, blue, "t^+"'] \& \color{blue} T^+
\end{tikzcd}
\end{equation}

Note that, by the pasting lemma for POs, since the phased rewrite of $G$, by $r'$ then $r^+$, occurs through consecutive POs, its overall effect is the same as that obtained by applying $r$ directly. Note also that we could instead have constructed $T^+$ by taking the PO of $r^+$ and $x = h' \circ m'$ and applying the UP of $G^+$ to construct $h^+$; the two approaches are equivalent by pasting for POs.

The propagated rewrite $t^+ : T \rightarrow T^+$ performs all the additions and merges, as specified by $r^+$ for $G'$, in $T$ to produce the new type $T^+$ required for $G^+$. We can alternatively obtain this rewrite by projecting $r^+$ to a new rule that applies specifically to $T$.

\paragraph{Definition} The \emph{projection} $\hat{r}^+ : L_T \rightarrow L_T^+$ of $r^+$ to $T$ is computed by taking the IF of $h' \circ m'$ followed by the PO of $r^+$ and $\hat{h}'$. It immediately has an expansive instance $\hat{m}'$ in $T$ and we obtain $T^+$ by taking the PO of $\hat{m}'$ and $\hat{r}^+$.
\begin{equation}\label{f-proj}
\begin{tikzcd}[ampersand replacement=\&]
	 L' \arrow[ddr, bend right, "h' \circ m'"'] \arrow[dr, green!50!black, "\hat{h}'"'] \arrow[r, "r^+"] \& L^+ \arrow[dr, blue]  \\
	 \& \color{green!50!black} L_T \arrow[r, blue, "\hat{r}^+"'] \arrow[d, green!50!black, tail, "\hat{m}'"'] \& L_T^+ \arrow[d, blue, tail, "\hat{m}^+"] \\
	 \& T \arrow[r, blue, "t^+"'] \& T^+
\end{tikzcd}
\end{equation}
The retyping of $G^+$ by $T^+$ follows from a straightforward application of the UP of the PO, in (\ref{f-canonical}), defining $G^+$.

It is easy to show, by the pasting lemma for POs, that these two definitions of $T^+$ coincide; as such, for the instance $h : G \rightarrow T$ of the simple hierarchy $e : d \rightarrow s$, we can use either. However, in a more general setting where $T$ may be typed by further objects, we must compute the rule projection explicitly in order to continue propagation; we return to this in section 5.

\subsection{The forward clean-up phase}

The strict phase of rewriting allows us to add elements to $G$ that can already be typed by $T$. However, if we wish to add elements that cannot be typed by $T$, this must occur during the canonical phase of rewriting; as such, every such element acquires a \emph{distinct} type in the updated $T^+$.

In order to allow the addition of multiple elements of the \emph{same} new type in $T^+$, we allow the specification of a \emph{clean-up} phase of rewriting, that applies only to $T^+$ (and not $G^+$), by providing an epi $r^\oplus : L_T^+ \twoheadrightarrow L_T^\oplus$; this allows us in particular to merge two newly-added elements of $T^+$. However, this requires us to know $L_T^+$---which is dependent on the typing $h' : G' \rightarrow T$ and so cannot be specified statically, at the same time as $r$, but rather dynamically when $r$'s rewrite is propagated to $T$.

\paragraph{Definition}
The clean-up phase is specified by an epi $r^\oplus : L_T^+ \twoheadrightarrow T^\oplus$ and the expansive instance $\hat{m}^+ : L_T^+ \rightarrowtail T^+$, obtained after the rewrite of $T$ with the rule projection $\hat{r}^+$ above, giving rise to the final retyping $t^\oplus \circ h^+ : G^+ \rightarrow T^\oplus$ of $G^+$.
\begin{equation}\label{f-cleanup}
\begin{tikzcd}[ampersand replacement=\&]
	G^+ \arrow[dr, "h^+"'] \& L_T^+ \arrow[d, tail, "\hat{m}^+"] \arrow[r, two heads, "r^\oplus"] \& L_T^\oplus \arrow[d, tail, "\hat{m}^\oplus"] \\
	 \& T^+ \arrow[r, two heads, "t^\oplus"'] \& T^\oplus
\end{tikzcd}
\end{equation}

Let us note that if $r'$ adds an element $e_1$ to $G$ and $r^+$ adds a second $e_2$, the clean-up phase may merge the newly-added element of $T^+$ with the element of $T$ that types $e_1$, so that $e_1$ and $e_2$ have the same type in $T^\oplus$. This enables us to correct the update in the case where, at the time of defining $r$ and its factorization, we failed to realize that $e_2$ could actually be typed in $T$.

We ask for $r^\oplus$ to be an epi because, in all concrete models of interest to us, this corresponds to a rule that \emph{only} merges nodes of $T^+$ in the image of $\hat{m}^+$ and we have no use case for using clean-up to add new elements to $T^+$.

\subsection{Example}

Let us illustrate the above theory in a case where $G$ and $T$ are sets, \ie the hierarchy represents a multi-set. The object $T$ has two elements---white and black circle---and $G$ has two instances of each (we use this colour coding to avoid specifying the homomorphisms explicitly). The rule specifies (i) the \emph{merge} of one white and one black circle; and (ii) the addition of two squares (which we wish to have the same type).
\[\includegraphics[scale=0.58]{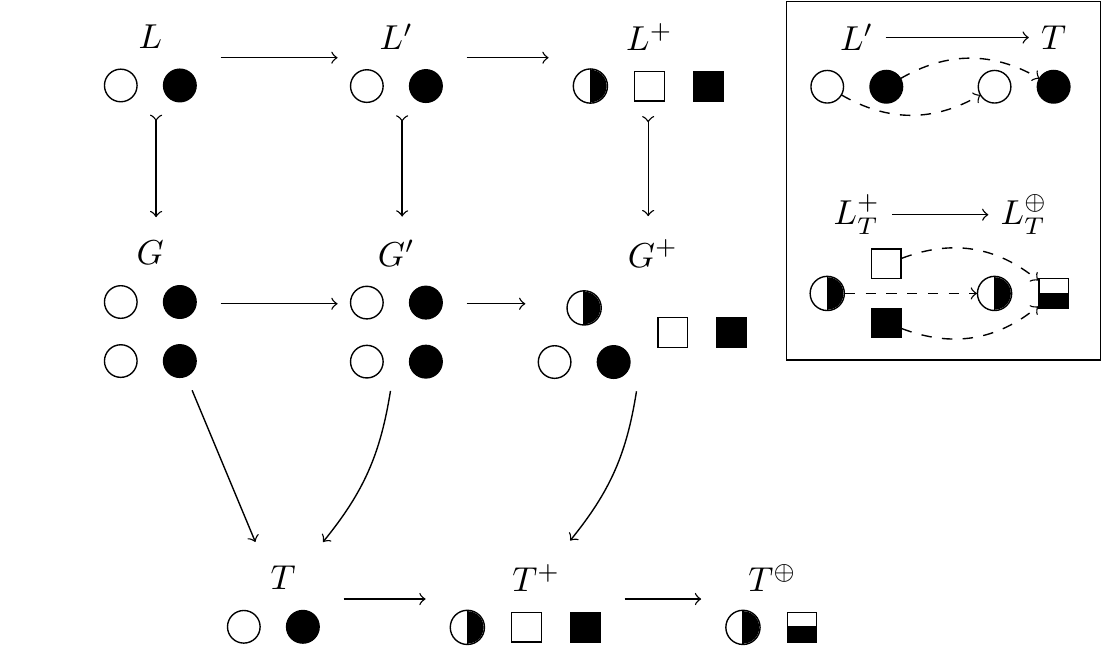}\]

The strict phase can neither merge the circles nor add the squares; as such, everything must occur in the canonical phase which performs, and propagates, the merge and additions. Note that this has the \emph{side-effect} that the two circles of $G$ not directly concerned by the rewrite have nonetheless been retyped in $T^+$. The clean-up phase now allows us to merge the two newly-added squares of $T^+$ so that the two squares in $G^+$ have the same type in $T^\oplus$.

If a square already exists in $T$, we could factorize the rule differently to add one square in the strict phase. In this case, clean-up can be used to merge the single newly-added square in $T^+$ with the one that existed in $T$; the overall effect is the same as if we had simply added both squares in the strict phase.
\[\includegraphics[scale=0.58]{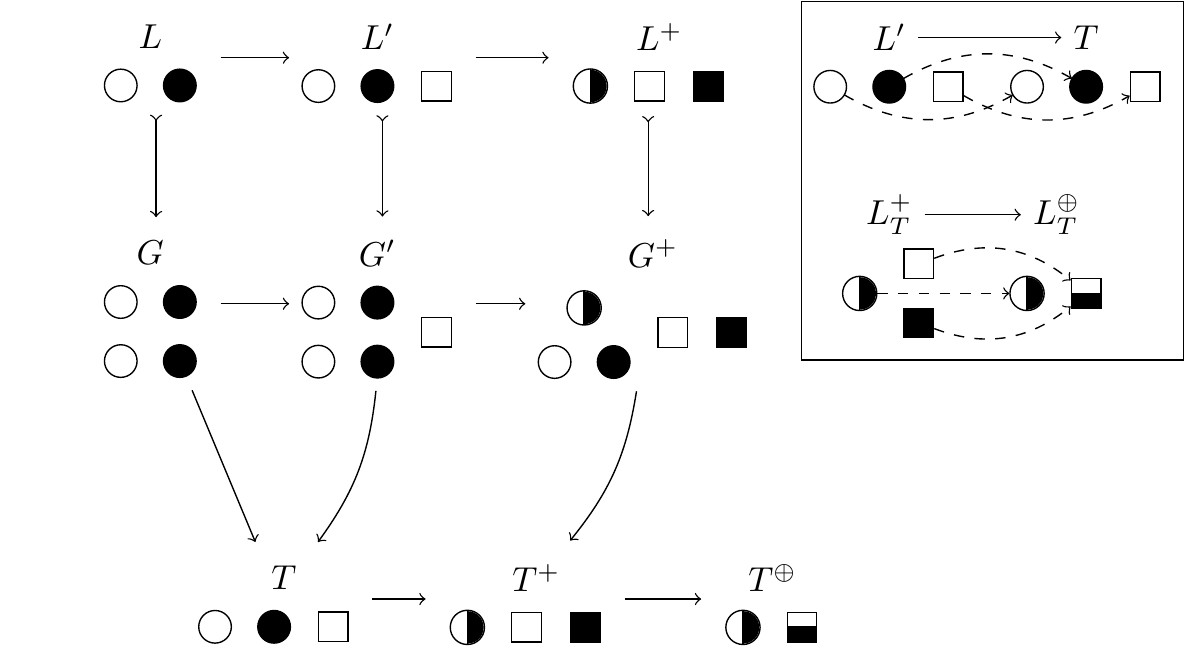}\]

\section{Backward propagation}

In this section, we consider a rule $r : L \leftarrow L^-$ with a restrictive instance $m : L \rightarrowtail T$ in $T$. We can immediately compute the PB of $h$ and $m$ to obtain a span $\hat{m} : G \leftarrowtail L_G \rightarrow L : \hat{h}$ from the object $L_G$ that can be seen as the sub-object of $G$ whose \emph{typing} by $T$ can be modified by $r$.
\begin{equation}\label{LG}
\begin{tikzcd}[ampersand replacement=\&]
	 \color{blue} L_G \arrow[d, blue, tail, "\hat{m}"'] \arrow[r, blue, "\hat{h}"] \& L \arrow[d, tail, "m"] \\
	 G \arrow[r, "h"'] \& T
\end{tikzcd}
\end{equation}

\subsection{The strict phase of backward rewriting}

Analogously to forward propagation, we must provide a factorization of $r$ in order to specify which changes to $T$ are to be propagated to $G$.

\paragraph{Definition}
Given a rule $r : L \leftarrow L^-$, a \emph{backward factorization} of $r$ is an object $L'$ and arrows $r' : L \leftarrow L'$ and $r^- : L' \leftarrow L^-$ such that $r = r^- \circ r'$; and an arrow $\hat{h}' : L_G \rightarrow L'$ such that $\hat{h} = r' \circ \hat{h}'$. Note that $L_G$ (not $G$) plays the role analogous to $T$ in forward propagation.
\begin{equation}\label{b-fact}
\begin{tikzcd}[ampersand replacement=\&]
	L \& L^- \arrow[l, "r"'] \arrow[d, red, "r^-"] \\ 
	L_G \arrow[u,"\hat{h}"] \arrow[r, red, "\hat{h}'"'] \& \color{red} L'  \arrow[ul, red, "r'" description]
\end{tikzcd}
\end{equation}

The factorization of $r$ splits its application into two phases: the \emph{strict} phase, specified by $r'$, which modifies only $T$; and the \emph{canonical} phase, specified by $r^-$, which modifies $G$ and $T$. As such, in the strict phase of restrictive rewriting, $G$ and $L_G$ remain invariant.

\paragraph{Definition}
The \emph{strict rewrite} of $T$ is defined by taking the PBC of $r'$ and $m$. By definition (\ref{b-fact}) and an application of the UP of this PBC to the PB, in (\ref{LG}), defining $L_G$, we obtain the retyping of $G$ as $h' : G \rightarrow T'$.
\begin{equation*}
\begin{tikzcd}[ampersand replacement=\&]
	 L \arrow[d, tail, "m"'] \& L' \arrow[l, "r'"'] \arrow[d, blue, tail, "m'"] \& \& \\
	 T \& \color{blue} T' \arrow[l, blue, "t'"] \&
\end{tikzcd}
\begin{tikzcd}[ampersand replacement=\&]
	 L_G \arrow[d, tail, "\hat{m}"'] \arrow[dr, "\hat{h}"'] \arrow[drr, "\hat{h}'"] \\
	 G \arrow[drr, dashed, "h'" description, pos=0.4] \arrow[dr, "h"'] \& L \arrow[d, tail] \& L' \arrow[l] \arrow[d, tail] \\
	 \& T \& T' \arrow[l]
\end{tikzcd}
\end{equation*}

Note that any element of $T$ that is deleted must have \emph{no instances} in $G$ for this to be possible---this is a consequence of the requirement that $\hat{h} = r' \circ \hat{h}'$; and that, if an element of $T$ is cloned, \emph{all} its instances in $G$ are reassigned a \emph{unique} type in $T'$ by $\hat{h}'$, \ie we are performing a \emph{concept refinement}.

Note also that, by the inverse pasting lemma for PBs, the resulting square is itself a PB:
\begin{equation}\label{b-retype}
\begin{tikzcd}[ampersand replacement=\&]
	 L_G \arrow[d, tail, "\hat{m}"'] \arrow[r, "\hat{h}'"] \& L' \arrow[d, tail, "m'"] \\
	 G \arrow[r, "h'"'] \& T'
\end{tikzcd}
\end{equation}

\subsection{The canonical phase of backward propagation}

\paragraph{Definition}
The \emph{rewrite} of $T$ is completed by taking the PBC of $r^-$ and $m'$. The \emph{backward propagation} to $G$ is then defined by taking the PB of $h'$ and $t^-$. The final typing of $G^-$ by $T^-$ is simply $h^-$.
\begin{equation}\label{b-canonical}
\begin{tikzcd}[ampersand replacement=\&]
	 L' \arrow[d, tail, "m'"'] \& L^- \arrow[l, "r^-"'] \arrow[d, blue, tail, "m^-"] \& \& G \arrow[d, "h'"'] \& \color{blue} G^- \arrow[l, blue, "g^-"'] \arrow[d, blue, "h^-"] \\
	 T' \& \color{blue} T^- \arrow[l, blue, "t^-"] \& \& T' \& T^- \arrow[l, "t^-"]
\end{tikzcd}
\end{equation}

This construction is analogous to the direct construction of $T^+$ as a PO in forward propagation. If the strict phase of rewriting is trivial, \ie $L \iso L'$, this corresponds exactly to the notion of (backward) propagation defined in \cite{harmer2017hdr}. Note that, by the horizontal pasting lemma for PBCs (see, for example, Proposition 5 of \cite{lowe2010graph}), the overall effect of $r'$ followed by $r^-$ on $T$ is the same as that obtained by applying $r$ directly.

The propagated rewrite $g^- : G \leftarrow G^-$ performs all the clones and deletions, as specified by $r^-$ for $T'$, in $G$ to produce the new object $G^-$ typed by $T^-$. We can also obtain this rewrite by constructing a new rule applying directly to $G$.

\paragraph{Definition}
The \emph{lifting} $\hat{r}^- : L_G \leftarrow L_G^-$ of $r^-$ to $G$ is computed by taking the PB of $\hat{h}'$ and $r^-$. It immediately has the restrictive instance $\hat{m}$, from (\ref{LG}),  in $G$ from which we obtain $G^-$ by taking the PBC of $\hat{r}^-$ and $\hat{m}$.
\begin{equation}\label{b-lift}
\begin{tikzcd}[ampersand replacement=\&]
 	L_G \arrow[d, "\hat{h}'"'] \& \color{blue} L_G^- \arrow[l, blue, "\hat{r}^-"'] \arrow[d, blue, "\hat{h}^-"] \& \& \\
	L' \& \arrow[l, "r^-"] L^-
\end{tikzcd}
\begin{tikzcd}[ampersand replacement=\&]
 	L_G \arrow[d, tail, "\hat{m}"'] \& \color{blue} L_G^- \arrow[l, "\hat{r}^-"'] \arrow[d, blue, tail, "\hat{m}^-"] \\
	G \& \arrow[l, blue, "g^-"] \color{blue} G^-
\end{tikzcd}
\end{equation}

In this case, we must construct the new typing of $G^-$ by $T^-$ by applying the pasting lemma for PBs and the UP of the PBC defining $T^-$, in (\ref{b-canonical}), to obtain $h^- : G^- \rightarrow T^-$.
\begin{equation}\label{b-reconstruct}
\begin{tikzcd}[ampersand replacement=\&]
	L' \arrow[d, "m'"', tail] \&  L_G \arrow[l, "\hat{h}'"'] \arrow[d, tail, "\hat{m}"] \& L^-_G \arrow[l, "\hat{r}^-"'] \arrow[d, tail, "\hat{m}^-"] \& \\
	T' \& G \arrow[l, "h'"] \& G^- \arrow[l, "g^-"] \&
\end{tikzcd}
\begin{tikzcd}[ampersand replacement=\&]
	 L_G^- \arrow[d, "\hat{m}^-"'] \arrow[dr, "\hat{h}' \circ \hat{r}^-" description] \arrow[drr, "\hat{h}^-"] \\
	 G^- \arrow[drr, dashed, "h^-" description, pos=0.4] \arrow[dr, "h' \circ g^-"'] \& L' \arrow[d, tail] \& L^- \arrow[l] \arrow[d, tail] \\
	 \& T' \& T^- \arrow[l]
\end{tikzcd}
\end{equation}


The proof of the equivalence of the two definitions of $G^-$ is a little more complex than its analogue for forward propagation; we give its proof here. We begin by constructing a commutative cube whose left face is the PB (\ref{b-retype}), whose front and bottom faces are respectively the PBC and PB of (\ref{b-canonical}) and whose top face is the PB of (\ref{b-lift}). By the universal property of $G^-$, there is a unique way to complete this to a commutative cube and, by inverse pasting for PBs, the back face (and indeed the right face) is a PB:
\begin{equation}\label{b-cube}
\begin{tikzcd}[row sep=small, column sep=small]
L_G \ar[dd, tail, "\hat{m}"'] \ar[dr, "\hat{h}'"'] & & \ar[ll, "\hat{r}^-"'] \ar[dd, tail, dotted, "z" near start] \ar[dr, "\hat{h}^-"] L_G^- \\
& L' & & L^- \ar[dd, tail, "m^-"] \ar[ll, crossing over, "r^-"' near end] \\
G \ar[dr, "h'"'] & & G^- \ar[ll, "g^-"' near start] \ar[dr, "h^-"] \\
& T' \ar[from=uu, tail, crossing over, "m'"' near start] & & T^- \ar[ll, "t^-"]
\end{tikzcd}
\end{equation}

\begin{Prop}
The back face of (\ref{b-cube}) is a PBC.
\end{Prop}
\begin{Pf}
Suppose we have a PB that factors through the back face:
\[\begin{tikzcd}[row sep=small, column sep=small]
& & & X \arrow[dlll, bend right, "x"'] \ar[dd, "f"] \ar[dl, "x^-" description] \ar[dlll, phantom, "="] \\
L_G \ar[dd, tail, "\hat{m}"'] && \ar[ll, "\hat{r}^-"'] \ar[dd, tail, "z"] L_G^- \\
& & & Y \ar[dlll, bend left=100, "y"] \\
G && G^- \ar[ll, "g^-"] \\
\end{tikzcd}\]
By pasting this PB with (\ref{b-retype}), we obtain a PB that factors through the front face and, by the UP of $T^-$, we have a unique arrow $y_{T^-} : Y \rightarrow T^-$ such that (i) $t^- \circ y_{T^-} = h' \circ y$ and (ii) $y_{T^-} \circ f = m^- \circ \hat{h}^- \circ x^-$.

By (i), we apply the UP of $G^-$ to obtain a unique arrow $y_{G^-} : Y \rightarrow G^-$ satisfying:
\[\begin{tikzcd}
& Y \arrow[d, dotted, "y_{G^-}" description] \ar[ddl, bend right, "y"'] \ar[ddr, bend left, "y_{T^-}"] \ar[ddr, phantom, "(iv)"] \ar[ddl, phantom, "(iii)"] \\
& G^- \ar[dl, "g^-"] \ar[dr, "h^-"'] \\
G \ar[dr, "h'"'] & & T^- \ar[dl, "t^-"] \\
& T'
\end{tikzcd}\]

By diagram chase, we can apply the UP of $G^-$ a second time to obtain a unique arrow $x_{G^-} : X \rightarrow G^-$ satisfying:
\[\begin{tikzcd}
& X \arrow[d, dotted, "x_{G^-}" description] \ar[ddl, bend right, "\hat{m} \,\circ\, x"'] \ar[ddr, bend left, "m^- \,\circ\,\hat{h}^- \,\circ\, x^-"] \ar[ddr, phantom, "(vi)"] \ar[ddl, phantom, "(v)"] \\
& G^- \ar[dl, "g^-"] \ar[dr, "h^-"'] \\
G \ar[dr, "h'"'] & & T^- \ar[dl, "t^-"] \\
& T'
\end{tikzcd}\]
We have two candidate arrows from $X$ to $G^-$: $z \circ x^-$ and $y_{G^-} \circ f$. The first satisfies (v) and (vi) immediately. The second satisfies (v) because $g^- \circ y_{G^-} \circ f = y \circ f$, by (iii), and $y \circ f = \hat{m} \circ x$; and satisfies (vi) because $h^- \circ y_{G^-} \circ f = y_{T^-} \circ f$, by (iv), and $y_{T^-} \circ f = m^- \circ \hat{h}^- \circ x^-$, by (ii).

As such, $y_{G^-}$ is the unique arrow satisfying (i) and (ii) as required.
\qed\end{Pf}

This establishes that the two definitions of $G^-$ coincide. The above argument amounts to a proof of the abstract property that PBCs are stable under PBs: this requires a commutative cube whose front face is a PBC, whose left and bottom faces are PBs and where any one of the other faces is also a PB (in our case, the top face); see also Proposition 6 of \cite{lowe2010graph} or Lemma 1 of \cite{danos2014reversible}.


\subsection{The backward clean-up phase}

We specify the clean-up phase by providing a mono $r^\ominus : L_G^- \leftarrow L_G^\ominus$. Clearly, and analogously to the situation for forward propagation, in order to provide such an $r^\ominus$, we already need to know $L_G^-$---which is dependent on the typing $G \rightarrow T'$. As such, $r^\ominus$ cannot be specified statically but should rather be provided dynamically at the time that $r$'s rewrite is being propagated to $G$.
\begin{equation}
\begin{tikzcd}[ampersand replacement=\&]
	\& L_G^- \arrow[d, "\hat{m}^-"', tail] \&  L_G^\ominus \arrow[l, tail, "r^\ominus"'] \arrow[d, tail, "\hat{m}^\ominus"] \\
	T^- \& G^- \arrow[l, "h^-"'] \& G^\ominus \arrow[l, tail, "g^\ominus"]
\end{tikzcd}
\end{equation}

The clean-up phase allows us to remove undesired element clones that were not specified during the strict phase of rewriting, \eg a \emph{partial} concept refinement where some instances of a cloned element cannot be assigned a unique type in $T'$. However, if $r^\ominus$ is not a mono, this phase can also create additional clones, beyond what was specified by $r$, and we have no use case for this extra generality, just as we have no use case for allowing the clean-up phase to add new elements to $T^+$ during forward propagation.

\subsection{Example}

We again consider an example on a multi-set. The rule specifies (i) the deletion of the circle; and (ii) the cloning of the square into a (white) square and a black square. The fact that one square in $G$ is to become white while the other becomes black is expressed by the arrow from $L_G$ to $L'$.
\[
\includegraphics[scale=0.55]{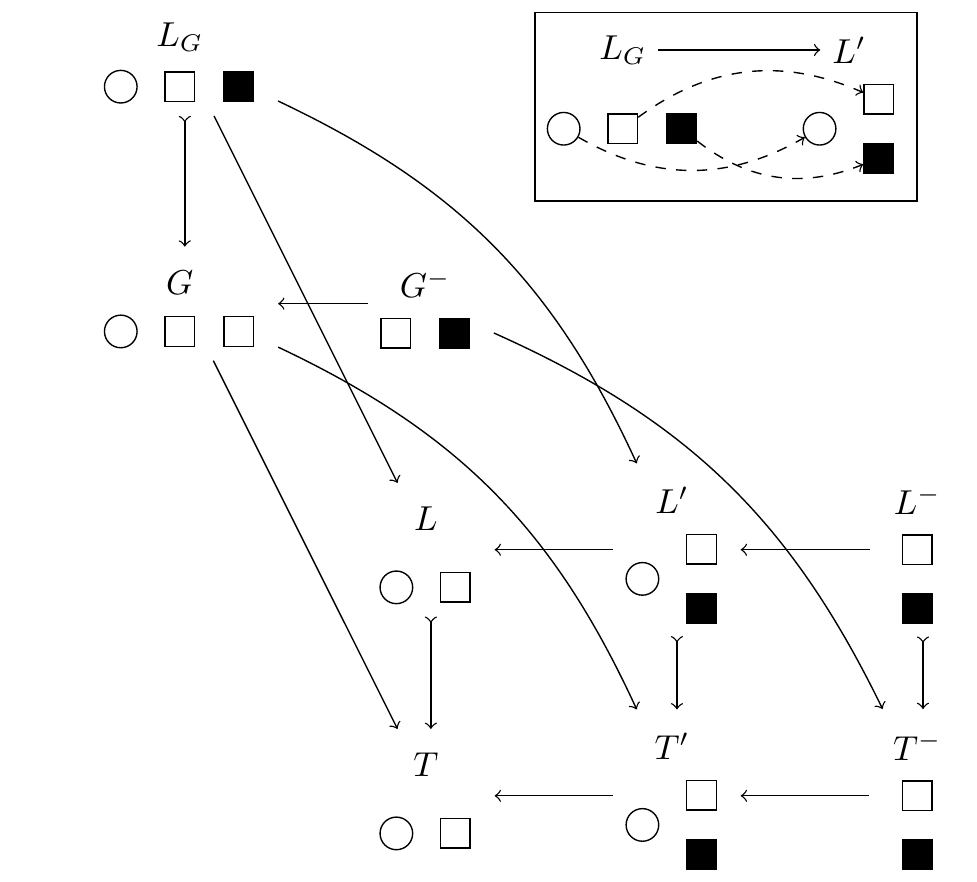}
\]
The strict phase of rewriting clones the square and retypes $G$, thus effecting a concept refinement; the canonical phase deletes the circle and propagates, thus deleting all circles in $G$.

If we have a third instance of the square in $G$ for which we \emph{cannot} assign a unique new type in $T'$, we must displace the cloning operation to the second phase of rewriting and propagate to \emph{all} instances of the square. In order to recover the same retyping of (the first two) squares as above, we must apply a clean-up rule to delete the unwanted clones.
\[
\includegraphics[scale=0.55]{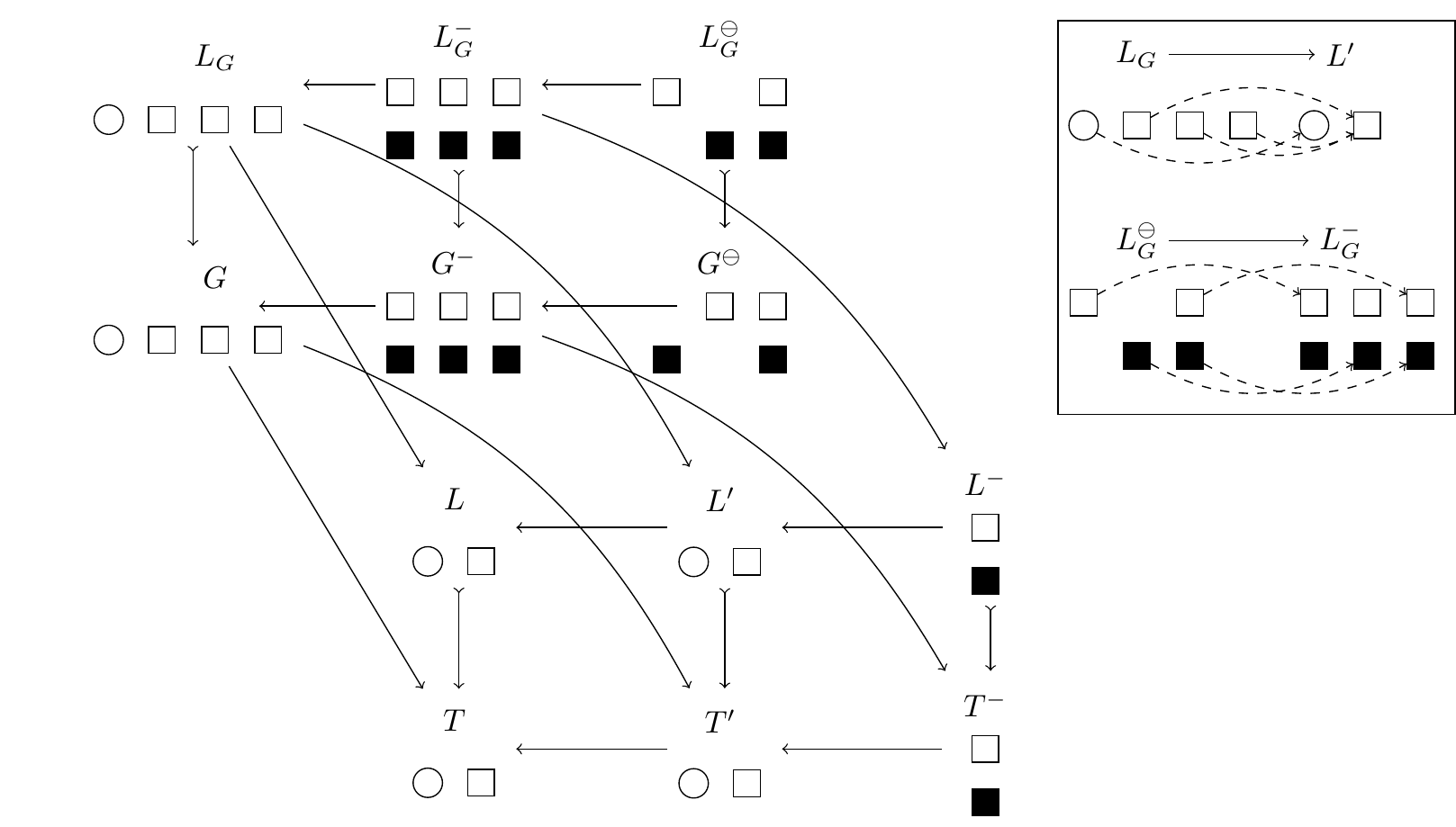}
\]

\section{Rewriting general hierarchies}

In this section, we consider the problem of how to update an instance of an arbitrary hierarchy and, in particular, how to do this \emph{in-place} so that the hierarchy remains valid at all times during the propagation process.

%
%

Given a DAG $H$ and a node $s$, we define the \emph{forward sub-graph} $\vec{H}_s$ to be the largest sub-graph of $H$ where $s$ is the unique source node. Dually, we define the \emph{backward sub-graph} $\cev{H}_s$ to be the largest sub-graph of $H$ where $s$ is the unique sink node.

In an instantiation of $H$ in a category $\C$, $\vec{H}_s$ identifies all the nodes $n$ of $H$ whose instantiations $\sem{n}$ can be affected, by propagation, by an expansive rewrite of the object $\sem{s}$. We need not propagate an expansive rewrite of $\sem{s}$ to an object $\sem{n}$ having an arrow \emph{into} $\sem{s}$ because it remains typed by post-composition with the arrow $\sem{s} \rightarrow \sem{s}^+$ induced by the rewrite.

Dually, $\cev{H}_s$ identifies all the nodes $n$ of $H$ that can be affected by a restrictive rewrite of $\sem{s}$: an object $\sem{n}$ having an arrow \emph{from} $\sem{s}$ still types the updated object $\sem{s}^-$ by pre-composition with the induced arrow $\sem{s} \leftarrow \sem{s}^-$.

\subsection{Expansive rewriting of a hierarchy}

In section 3, we have seen how the expansive rewrite of an object $G$, typed by a second object $T$, is factorized into a strict update, producing a $G'$ still typed by $T$, followed by a canonical update that produces a $G^+$ and propagates to $T$, yielding an updated $T^+$ that types $G^+$. However, in a general hierarchy, two orthogonal complications may arise: firstly, $G$ may be typed by several graphs and their respective factorizations may be incompatible; and secondly, we may have chains of typing $G_0 \rightarrow G_1 \rightarrow \cdots G_n$ for which we need to verify the compatability of the factorizations.

To illustrate the first point, let us consider the hierarchy $n_1 \leftarrow n_0 \rightarrow n_2$ instantiated in $\Set$ as follows: $\sem{n_0} = \emptyset$, $\sem{n_1} = \set{\circ}$ and $\sem{n_2} = \set{\bullet}$. If we update $\sem{n_0}$ with the rule $r : \emptyset \rightarrow \set{\circ,\bullet}$, the factorization with respect to $\sem{n_1}$ should specify that we add $\circ$ to $\sem{n_0}$ by a strict update and then add $\bullet$ and propagate this to $\sem{n_1}$; for $\sem{n_2}$, we need to do the opposite. As such, either strict update breaks one of the typing arrows. The natural solution is to propagate first, adding $\bullet$ to $\sem{n_1}$ and $\circ$ to $\sem{n_2}$, and then apply the entire rule $r$ to $\sem{n_0}$ as a \emph{strict} update---which is now possible, as an in-place update, because we have already updated $\sem{n_1}$ and $\sem{n_2}$.

For the second point, consider an instantiated hierarchy $G_0 \rightarrow G_1 \rightarrow G_2$, a rule $r : L \rightarrow L^+$ with an expansive instance $m : L \rightarrowtail G_0$ and factorizations $r_1, r_1^+, x_1$ and $r_2, r_2^+, x_2$ through $L_1$ and $L_2$ for $G_1$ and $G_2$ respectively.
\[\begin{tikzcd}
& \\
L \arrow[d,tail] \arrow[rr, bend left, "r_2" description] \arrow[r, "r_1" description] & L_1 \arrow[ddl, "x_1" description] \arrow[rr, bend right, "r_1^+"'] \arrow[dl, phantom, "=", near end] & L_2 \arrow[r, "r_2^+"] \arrow[dddll, "x_2" description] \arrow[ddll, phantom, "="] & L^+ \\
G_0 \arrow[d] \\
G_1 \arrow[d] \\
G_2
\end{tikzcd}\]
If $r_1$ adds an element to $G_0$, necessarily typed in $G_1$ and so in $G_2$ as well, but $r_2$ postpones this operation to $r^+$, therefore typing the element in $G_2^+$ but not $G_2$, it will not be possible to type $G_1^+$ by $G_2^+$. We must therefore require that $r_2$ extends $r_1$; we formalize this with the so-called \emph{composability} condition.


Suppose we have an instantiated hierarchy $\sem{H}$ containing an object $G_0 = \sem{n_0}$ and a rule $r : L \rightarrow L^+$ with an expansive instance $m : L \rightarrowtail G_0$. For any object $G_i \in \sem{\vec{H}_{n_0}}$, define $h_i$ to be the unique homomorphism from $G_0$ to $G_i$ obtained by composing any path from $G_0$ to $G_i$. (They are all equal by the commutativity condition.)

\paragraph{Definition}
Given an arrow $h_{ij} : G_i \rightarrow G_j$ in $\sem{\vec{H}_{n_0}}$ and factorizations
\begin{equation}\label{f-facts}
\begin{tikzcd}[ampersand replacement=\&]
	L \arrow[d, "h_i \circ m"'] \arrow[r, "r"] \arrow[dr, red, "r_i" description] \& L^+ \& \& L \arrow[d, "h_j \circ m"'] \arrow[r, "r"] \arrow[dr, red, "r_j" description] \& L^+ \\
	G_i \& \color{red} L_i \arrow[u, red, "r_i^+"'] \arrow[l, red, "x_i"] \& \& G_j \& \color{red} L_j \arrow[u, red, "r_j^+"'] \arrow[l, red, "x_j"]
\end{tikzcd}
\end{equation}%
that define the propagation of $r$ to $G_i$ and $G_j$ respectively, we say that these factorizations are \emph{composable} iff there exists an arrow $\ell_{ij} : L_i \rightarrow L_j$ satisfying:
\begin{equation}\label{f-ell}
	\begin{tikzcd}[ampersand replacement=\&, column sep=small, row sep=small]
		\& L_i  \arrow[dd, red, "\ell_{ij}" description] \arrow[dr, "r_i^+"] \& \\
		L \arrow[rr, phantom, "=" near start] \arrow[rr, phantom, "=" near end] \arrow[ur, "r_i"] \arrow[dr, "r_j"'] \& \& L^+ \& \& \\
		\& L_j \arrow[ur, "r_j^+"'] \&
	\end{tikzcd}
	\begin{tikzcd}[ampersand replacement=\&, row sep=small]
		L_i \arrow[dd, "x_i"'] \arrow[r, red, "\ell_{ij}"] \arrow[ddr, phantom, "="] \& L_j \arrow[dd, "x_j"] \\
		\\
		G_i \arrow[r, "h_{ij}"'] \& G_j
	\end{tikzcd}
\end{equation}

In the case above, this enables us to update $G_2$, by taking the PO of $x_2$ and $r_2^+$, and then $G_1$, by taking the PO of $x_1$ and $r_1^+ = r_2^+ \circ \ell_{12}$. An immediate application of the UP of $G_1^+$, using (\ref{f-ell}), gives us the new typing $h_{12}^+ : G_1^+ \rightarrow G_2^+$.
\[\begin{tikzcd}
& L_1 \arrow[ddl, "x_1" description] \arrow[rr, bend left, "r_1^+"] \arrow[r, "\ell_{12}"] & L_2 \arrow[r, "r_2^+"] \arrow[dddll, "x_2" description] & L^+ \arrow[ddd, bend left] \arrow[dd] \\
\\
G_1 \arrow[d, "h_{12}"'] \arrow[rrr] & & & G_1^+ \arrow[d, dotted, "h_{12}^+"'] \\
G_2 \arrow[rrr] & & & G_2^+
\end{tikzcd}\]
Note that the update of $G_1$ is morally a \emph{strict} update using the rule $r_1^+$. As written, this is not a \emph{bona fide} rule application because $x_1$ is not necessarily a mono; however, this is equivalent to applying the projection of $r_1$ as explained in section 3.
We conclude by updating $G_0$, by taking the PO of $m$ and $r$, and obtain the retyping $h_{01}^+ : G_0^+ \rightarrow G_1^+$ as per (\ref{f-strict}).


In general, suppose we have factorizations $r_i : L \rightarrow L_i$, $r_i^+ : L_i \rightarrow L^+$ and $x_i : L_i \rightarrow G_i$ for all $G_i \in \sem{\vec{H}_{n_0}}$ ($i \neq 0$) and, for each $h_{ij} : G_i \rightarrow G_j \in \sem{\vec{H}_{n_0}}$, an arrow $\ell_{ij} : L_i \rightarrow L_j$ satisfying (\ref{f-ell}). We wish to define the updated hierarchy $\sem{\vec{H}_{n_0}}^+$. If $\sem{\vec{H}_{n_0}}$ is just $G_0$, we rewrite it to $G_0^+$ and $\sem{\vec{H}_{n_0}}^+$ is trivially valid. Otherwise, we define $\sem{\vec{H}_{n_0}}^+$ as follows:
\begin{itemize}
\item
We update each sink node $G_s$ to $G_s^+$, by taking the PO of $x_s$ and $r_s^+$.
\item
We consider the nodes $G_k$ that are sink nodes if we remove all the $G_s$s. We update each $G_k$, by taking the PO of $x_k$ and $r_k^+$, and apply the UP of each $G_k^+$, using (\ref{f-ell}), to update each arrow $h_{ks} : G_k \rightarrow G_s$ to $h_{ks}^+ : G_k^+ \rightarrow G_s^+$. We then continue inductively.
\end{itemize}
This procedure updates every node and every arrow of $\sem{\vec{H}_{n_0}}$, \ie it preserves the structure of the hierarchy.
\begin{Prop}
$\sem{\vec{H}_{n_0}}^+$ is a valid hierarchy.
\end{Prop}
\begin{Pf}
If $\sem{\vec{H}_{n_0}}^+$ is a tree, the result is immediate; so we only need to check that $\sem{\vec{H}_{n_0}}^+$ satisfies the commutativity condition. If there are multiple paths in $\sem{\vec{H}_{n_0}}^+$ from $G_i^+$ to $G_j^+$, each one comes from a distinct path from $G_i$ to $G_j$ in $\sem{\vec{H}_{n_0}}$. Each such path $G_i \rightarrow G_{i_1} \rightarrow \cdots G_{i_n} \rightarrow G_j$ gives rise to a factorization of $r_i^+$ as $r_j^+ \circ \ell_{i_nj} \circ \cdots \ell_{ii_1}$, by (\ref{f-ell}), and all of those paths are equal in $\sem{\vec{H}_{n_0}}$, by commutativity. Let us denote this path $p_{ij} : G_i \rightarrow G_j$.

By diagram chase, $\hat{r}_j^+ \circ p_{ij} \circ x_i = \hat{x}_j \circ r_i^+$ so, by the UP of $G_i^+$, we have a unique arrow $h_{ij}^+ : G_i^+ \rightarrow G_j^+$ satisfying $\hat{r}_j^+ \circ p_{ij} = h_{ij}^+ \circ \hat{r}_i^+$ and $\hat{x}_j = h_{ij}^+ \circ \hat{x}_i$. Each path $G_i \rightarrow G_{i_1} \rightarrow \cdots G_{i_n} \rightarrow G_j$ in $\sem{\vec{H}_{n_0}}$ gives rise to a path $G_i^+ \rightarrow G_{i_1}^+ \rightarrow \cdots G_{i_n}^+ \rightarrow G_j^+$ that satisfies this UP and so $\sem{\vec{H}_{n_0}}^+$ satisfies commutativity.
\qed\end{Pf}

Note that the structure of $\vec{H}_{n_0}$ imposes constraints on the order in which we update the associated objects: this guarantees that, at all times during update, the entire hierarchy remains in a valid state, \ie we can perform in-place update to obtain finally $\sem{H}^+$, \ie $\sem{H}$ where the sub-graph $\sem{\vec{H}_{n_0}}$ is updated to $\sem{\vec{H}_{n_0}^+}$.

In the simple situation studied in section 3, clean-up can be considered as a final step of the rewrite which applies only to $T$ and therefore cannot propagate. In a general hierarchy, a clean-up rule may itself further propagate and this requires us to specify factorizations and composability exactly as for $r$. As such, for the sake of avoiding redundancy, we consider it as a separate rule application and correctness follows by the above argument.

\subsection{Restrictive rewriting of a hierarchy}

Dually to the case of expansive rewriting, two complications may arise when we perform a restrictive update of an object $G_0$ in a general hierarchy: firstly, $G_0$ may type several objects and their respective factorizations may be incompatible; and secondly, we need to verify the compatibility of factorizations in chains of the form $G_n \rightarrow \cdots G_1 \rightarrow G_0$.

The first point typically arises in a hierarchy such as $G_1 \rightarrow G_0 \leftarrow G_2$ when we clone two nodes of $G_0$ and propagate one of the clones to $G_1$ and the other to $G_2$. As for the analogous case in expansive rewriting discussed above, this requires us to update $G_1$ and $G_2$ before $G_0$. However, in the case of restrictive rewriting, this forces us to use the lifting of the rule explicitly---because the direct construction of $G_1^-$ and $G_2^-$ uses $G_0^-$.

The second point requires us to specify composability conditions, analogous to those for expansive rewriting, in order to obtain a multi-stage factorization of $r$ for each chain of the hierarchy.


Suppose we have an instantiated hierarchy $\sem{H}$ containing an object $G_0 = \sem{n_0}$ and a rule $r : L \leftarrow L^-$ with a restrictive instance $m : L \rightarrowtail G_0$. For any object $G_i$ in the backward sub-graph $\cev{H}_{n_0}$, let $h_i$ be the unique homomorphism from $G_i$ to $G_0$. Given an arrow $h_{ij} : G_i \rightarrow G_j$ in $\sem{\cev{H}_{n_0}}$, we define $L_{G_i}$ and $L_{G_j}$ as in (\ref{LG}) and apply the UP of $L_{G_j}$ to obtain the unique arrow $\hat{h}_{ij} : L_{G_i} \rightarrow L_{G_j}$ satisfying:
\begin{equation}\label{b-rule-hier1}
\begin{tikzcd}
L_{G_i} \arrow[d, tail, "\hat{m}_i"'] \arrow[rr, bend left, "\hat{h}_i"] \arrow[dr, phantom, "="] \arrow[rr, bend left=15, phantom, "=" description] \arrow[r, dotted, "\hat{h}_{ij}"] & L_{G_j} \arrow[d, tail, "\hat{m}_j"] \arrow[r, "\hat{h}_j"] & L \arrow[d,tail, "m"] \\
G_i \arrow[r, "h_{ij}"'] \arrow[rr, bend right, "h_i"'] & G_j \arrow[r, "h_j"'] & G_0
\end{tikzcd}\end{equation}
The commuting triangle enables us to apply inverse pasting to establish that the commuting square is in fact a PB.

\paragraph{Definition}
Given an arrow $h_{ij} : G_i \rightarrow G_j$ in $\cev{H}_{n_0}$ and factorizations
\begin{equation}\label{b-facts}
\begin{tikzcd}[ampersand replacement=\&]
	L \& L^- \arrow[l, "r"'] \arrow[d, red, "r_i^-"] \& \\ 
	L_{G_i} \arrow[u,"\hat{h}_i"] \arrow[r, red, "\hat{h}'_i"'] \& \color{red} L_i  \arrow[ul, red, "r_i" description] \&
\end{tikzcd}
\begin{tikzcd}[ampersand replacement=\&]
	L \& L^- \arrow[l, "r"'] \arrow[d, red, "r_j^-"] \\ 
	L_{G_j} \arrow[u,"\hat{h}_j"] \arrow[r, red, "\hat{h}'_j"'] \& \color{red} L_j \arrow[ul, red, "r_j" description]
\end{tikzcd}
\end{equation}
that define the propagation of $r$ to $G_i$ and $G_j$ respectively, these factorizations are \emph{composable} iff there exists an arrow $\ell_{ij} : L_i \rightarrow L_j$ such that
\begin{equation}\label{b-ell}
	\begin{tikzcd}[ampersand replacement=\&, column sep=small, row sep=small]
		\& L_i \arrow[dl, "r_i"'] \arrow[dd, dashed, "\ell_{ij}" description] \\
		L \& \& L^- \arrow[ul, "r_i^-"'] \arrow[dl, "r_j^-"] \arrow[ll, phantom, "=" near start, "=" near end] \& \& \\
		\& L_j \arrow[ul, "r_j"]
	\end{tikzcd}
	\begin{tikzcd}[ampersand replacement=\&, row sep=small]
		L_i \arrow[dd, dashed, "\ell_{ij}"'] \& \arrow[l, "\hat{h}'_i"'] L_{G_i} \arrow[dd, "\hat{h}_{ij}"] \\
		\\
		L_j \& \arrow[l, "\hat{h}'_j"] L_{G_j}  \arrow[uul, phantom, "="]
	\end{tikzcd}
\end{equation}

For each $G_i$, the lifting $\hat{r}_i : L_{G_i} \leftarrow L_{G_i}^-$ of $r$ is defined by the PB of $\hat{h}'_i$ and $r_i^-$, as in (\ref{b-lift}), with instance $\hat{m}_i : L_{G_i} \rightarrowtail G_i$.

For each $h_{ij}$, we can connect the liftings of $\hat{r}_i$ and $\hat{r}_j$ by applying the UP of $L_{G_j}^-$, using (\ref{b-ell}), to obtain a unique arrow $\hat{h}_{ij}^- : L_{G_i}^- \rightarrow L_{G_j}^-$ satisfying:
\begin{equation}\label{b-rule-hier2}
\begin{tikzcd}
& L_{G_i} \arrow[dd, "\hat{h}'_i"' near start] \arrow[dr, phantom, "="] \arrow[dl, "\hat{h}_{ij}"'] & L_{G_i}^- \arrow[dd, bend left=50, "\hat{h}_i^-"] \arrow[l, "\hat{r}_i"] \arrow[d, dotted, "\hat{h}_{ij}^-"] \arrow[dd, bend left=30, phantom, "="] \\
L_{G_j} \arrow[d, "\hat{h}'_j"'] & & L_{G_j}^- \arrow[d, "\hat{h}_j^-"] \arrow[ll, "\hat{r}_j" near end] \\
L_j & L_i \arrow[l, "\ell_{ij}"] & L^- \arrow[l, "r_i^-"] \arrow[ll, bend left, "r_j^-"]
\end{tikzcd}
\end{equation}

If we have factorizations $r_i : L \leftarrow L_i$, $r_i^- : L_i \leftarrow L^-$ and $\hat{h}'_i : L_{G_i} \rightarrow L_i$ for all $G_i \in \sem{\cev{H}_{n_0}}$ ($i \neq 0$) and, for each $h_{ij} : G_i \rightarrow G_j \in \sem{\vec{H}_{n_0}}$, an arrow $\ell_{ij} : L_i \rightarrow L_j$ satisfying (\ref{b-ell}), we can now define the updated hierarchy $\sem{\cev{H}_{n_0}}^-$. If $\sem{\cev{H}_{n_0}}$ is just $G_0$, we rewrite it to $G_0^-$ and $\sem{\cev{H}_{n_0}}^-$ is trivially valid. Otherwise, we define it as follows:
\begin{itemize}
\item
We update each source node $G_s$ to $G_s^-$, by taking the PBC of $\hat{r}_s$ and $\hat{m}_i$.
\item
We consider the nodes $G_k$ that are source nodes if we remove all the $G_s$s. We update each $G_k$, by taking the PBC of $\hat{r}_k$ and $\hat{m}_k$, and apply the UP of each $G_k^+$, using (\ref{b-ell}), (\ref{b-rule-hier1}) and (\ref{b-rule-hier2}) as in (\ref{b-reconstruct}), to update each arrow $h_{sk} : G_s \rightarrow G_k$ to $h_{sk}^- : G_s^- \rightarrow G_k^-$. We then continue inductively.
\end{itemize}
This procedure updates every node and every arrow of $\sem{\cev{H}_{n_0}}$, \ie it preserves the structure of the hierarchy. The following diagram shows the general situation for an arrow $h_{ij}$.

\begin{equation}\label{b-recon}
\begin{tikzcd}
& L_{G_i} \arrow[dl, tail, "\hat{m}_i"] \arrow[dd, "\hat{h}_{ij}" near end] \arrow[ddddrr] & & & L_{G_i}^- \arrow[lll, "\hat{r}_i"] \arrow[dl, tail, "\hat{m}_i^-"] \arrow[dd, "\hat{h}_{ij}^-"] \\
G_i \arrow[dd, "h_{ij}"'] & & & G_i^- \arrow[lll, "g_i^-"' near start] \arrow[dd, dotted, "h_{ij}^-" near start] \\
& L_{G_j} \arrow[dl, tail, "\hat{m}_j"] \arrow[dd, "\hat{h}_j" near end] \arrow[ddr] & & & L_{G_j}^- \arrow[lll, "\hat{r}_j"] \arrow[dl, tail, "\hat{m}_j^-"] \arrow[dd, "\hat{h}_j^-"] \\
G_j \arrow[dd, "h_j"'] & & & G_j^- \arrow[lll, "g_j^-"' near start] \\
& L \arrow[dl, tail, "m"'] & \arrow[l] L_j & L_i \arrow[l] & L^- \arrow[l] \\
G_0
\end{tikzcd}
\end{equation}

\begin{Prop}
$\sem{\vec{H}_{n_0}}^-$ is a valid hierarchy.
\end{Prop}
\begin{Pf}
We only need to check the commutativity condition. Each path in $\sem{\cev{H}_{n_0}}^+$ from $G_i^-$ to $G_j^-$ comes from a distinct path from $G_i$ to $G_j$ in $\sem{\cev{H}_{n_0}}$. Each such path $G_i \rightarrow G_{i_1} \rightarrow \cdots G_{i_n} \rightarrow G_j$ gives rise to a factorization of $r_j^-$ as $\ell_{i_nj} \circ \cdots \ell_{ii_1} \circ r_i^-$, by (\ref{b-ell}), and all of those paths are equal in $\sem{\vec{H}_{n_0}}$, by commutativity. Let us denote this path $p_{ij} : G_i \rightarrow G_j$.

By the UP of $G_j^-$, as in (\ref{b-recon}), we have a unique arrow $h_{ij}^- : G_i^- \rightarrow G_j^-$ satisfying $p_{ij} \circ g_i^- = g_j^- \circ h_{ij}^- $ and $h_{ij}^- \circ \hat{m}_i^- = \hat{m}_j^- \circ \hat{h}_{ij}^-$. Each path $G_i \rightarrow G_{i_1} \rightarrow \cdots G_{i_n} \rightarrow G_j$ in $\sem{\cev{H}_{n_0}}$ gives rise to a path $G_i^- \rightarrow G_{i_1}^- \rightarrow \cdots G_{i_n}^- \rightarrow G_j^-$ that satisfies this UP and so $\sem{\cev{H}_{n_0}}^-$ satisfies commutativity.
\qed\end{Pf}

As for expansive rewriting of a hierarchy, the structure of $\cev{H}_{n_0}$ constrains the order in which we update to guarantee that, at all times during update, the entire hierarchy remains in a valid state, \ie we can perform in-place update to obtain finally $\sem{H}^-$, \ie $\sem{H}$ where the sub-graph $\sem{\cev{H}_{n_0}}$ is updated to $\sem{\cev{H}_{n_0}^+}$.

\section{Implementation and use cases}

\subsection{The \texttt{ReGraph} Python library}

In the preceding sections, we have detailed the mathematical theory of SqPO rewriting in general hierarchies. We have implemented this theory---at the time of writing, for the setting of \emph{simple} directed graphs with attributes on nodes and edges, although there is no conceptual or technological problem to extend this to directed \emph{multi}-graphs---in the \texttt{ReGraph} Python library\footnote{\texttt{https://github.com/Kappa-Dev/ReGraph}}. The library provides two back-ends: in-memory graphs, based on the \texttt{networkX} library widely used in complex systems---and persistent graphs using the Neo4j graph DB. Rules can be expressed declaratively, essentially using the mathematical definition used in this paper, or procedurally, using a simple language to express the primitive operations of clone, delete, add and merge.

Hierarchies, rewriting and propagation are implemented natively, in Python, for the in-memory back-end. However, the implementation of the persistent back-end requires a considerably greater effort because (i) Neo4j currently only provides a \emph{single} graph within which we must encode an arbitrary hierarchy; and (ii) all operations to be performed on that graph must be expressed through the Cypher query language used by Neo4j. The precise details of the encoding and the translation of rules into Cypher are highly technical and not of great interest in their own right. However, their ultimate effect is to enforce an \emph{abstraction barrier} that gives the illusion that Neo4j actually provides an implementation of our theory and, provided that the user only accesses Neo4j through \texttt{ReGraph}, guarantees that the current contents of the single underlying Neo4j graph always corresponds to a valid encoding of a hierarchy.

The principal difference between the theory presented in this paper and the implementation lies in the specification of propagation: in \texttt{ReGraph}, a controlled propagation is specified by a single relation that plays the same r\^ole as the strict and clean-up phases presented here. For example, the partial concept refinement of section 4.5 is expressed by the same rule together with a relation that specifies, for the first two squares, how to retype them; nothing need be specified for the third square which, as a result, is cloned.

\subsection{Graph databases}


Most modern graph DBs, such as Neo4j, are based on the notion of property graphs, \ie directed (multi-)graphs where nodes and edges may have a dictionary of \emph{properties} consisting of \emph{key-value} pairs. If we instantiate the above hierarchy in the category of property graphs, we obtain a setting where the graph $T$ defines the permitted node and edge types as well as all possible properties, \ie $T$ acts as a \emph{schema}. The graph $G$ defines the \emph{data} graph and the homomorphism to $T$ specifies the types of all its nodes and edges and guarantees that all edges and properties \emph{validate} the schema.

The generic implementation of \texttt{ReGraph} with persistent back-end thus provides, through the use of the $h : G \rightarrow T$ hierarchy, the illusion of a notion of schema for Neo4j graphs. In fact, we have also made an optimized implementation for this particular hierarchy which therefore avoids most of the overheads associated with the encoding into the single underlying Neo4j graph\footnote{\texttt{https://github.com/Kappa-Dev/ReGraph/blob/master/regraph/neo4j/graphs.py}}. In order to build a fully general front-end to Neo4j in this way, we need to extend \texttt{ReGraph} to work with non-simple graphs; we plan to do this in the near future. However, our work can also been viewed as a proposal for how to incorporate schema, or indeed mutliple graphs, \emph{natively} within a graph DB such as Neo4j rather than merely a means of encoding this. Indeed, the recently-launched ISO standardization process for GQL, and the associated informal working group PGSWG, is investigating ways to express multiple graphs and/or schema graphs. The extent to which our ideas are ultimately reflected in GQL will determine the extent to which our current encoding and implementation can be simplified and rendered more efficient.

In this setting, forward propagation constructs and applies an automatic update of the schema graph in the light of an update of the data graph that would otherwise have broken schema validation, \ie a descriptive update. Such updates typically occur during the earlier phases of application development where we do not yet have a clear picture of the structure of all the relevant data and therefore wish to allow the schema to evolve dynamically to accommodate the incoming data. Dually, backward propagation constructs an automatic update of the data graph in the event of an update of the schema, \ie a prescriptive update. Such updates more usually occur later in the development process where we wish to engineer specific refinements to the schema in the light of the needs of the application. A more detailed discussion can be found in \cite{bonifati2019schema}.

\subsection{Multi-set rewriting}

The simplest non-trivial hierarchy consists of two objects connected by a single arrow. Note that, in this case, the hierachy coincides with its skeleton. The instantiation of this hierarchy in $\mathbf{Set}$ specifies a set $T$ of \emph{types} and a set $G$ of \emph{instances} of these types where the typing of the elements of $T$ is given by the function from $G$ to $T$; as such, this provides an intensional representation of a multi-set over the set $T$.

The usual notion of multi-set rewriting operates only on $G$---the set $T$ is fixed in advance---but our framework provides a rigorous framework within which rewrites can also apply to $T$---either directly or, in practice more likely, through the forward propagation of a rewrite of $G$. In this way, the set of types can grow automatically, on the fly, an approach that otherwise requires a substantial algorithmic and implementation effort. However, for our implementation to provide an efficient simulation engine for such systems, further development would be necessary in order to exploit the intrinsic \emph{causality} between rules---which possible rule applications are created and destroyed by the application of a given rule---in order to maintain and update incrementally and efficiently the collection of all possible rule applications at any given time.

\subsection{The \texttt{KAMI} bio-curation system}

The original motivating use case for the development of the theory presented in this paper was the \texttt{KAMI} bio-curation system which provides a graph-based KR for protein-protein interactions (PPIs) in cellular signalling. The skeleton of \texttt{KAMI}'s hierarchy consists of a chain of three nodes and two edges: $N \rightarrow A \rightarrow M$ where $M$ represents the \emph{meta-model}, a graph defining the basic concepts of the system such as proteins, binding sites and binding interactions; $A$ represents the \emph{action graph}, which defines the particular proteins and interactions in a knowledge corpus; and $N$ represents the \emph{nuggets}, small graphs that capture the necessary conditions for a PPI to occur. A typical \texttt{KAMI} hierarchy has multiple nuggets but a single action graph and meta-model.

Unlike the previous use case, for graph DBs, the hierarchy itself can evolve over time: this usually occurs upon the addition of a completely new nugget but can also arise if a nugget is deleted or two nuggets are merged together (or even if a nugget is cloned although we have not yet needed to consider this case in practice). Such updates operate on the \emph{structure} of the KR rather than on its \emph{contents} and therefore lie out of the scope of this paper. However, we plan to investigate the nature of such updates, in the general context of graph hierarchies, as the nature of graph DBs containing multiple graphs remains a somewhat controversial issue for which no general consensus has yet been reached in the community.

The meta-model of \texttt{KAMI} is required to remain invariant under all (content) update operations. In other words, any update that propagates to the meta-model must be specified as being strict. In practice, this simply means that all new elements being added to a nugget must have a well-defined type in the meta-model---even if they do not yet exist in the action graph---and the update thus propagates, as necessary, to the action graph but no further.

The bio-curation tool \texttt{KAMI}\footnote{\texttt{https://github.com/Kappa-Dev/KAMI}} \cite{harmer2019bio,harmer2019kamistudio}, discussed in the introduction, is based on the \texttt{ReGraph} library. It makes extensive use of forward propagation, in order to aggregate new PPIs appropriately into an existing knowledge corpus, \eg if it identifies that a node mentioned in an input already exists in the action graph, it constructs a strict rewrite, to reuse that node, rather than creating a new one by canonical propagation. It also makes use of backward propagation in order to contextualize knowledge to a particular collection of gene products. Indeed, these were the original, informal use cases of propagation which motivated the development of the theory presented in this paper.

\section{Conclusions}

We have presented a formalism for graph-based knowledge representation and update that exploits SqPO rewriting to perform updates anywhere in a hierarchy of objects (typically sets or graphs). For this extended abstract, we have chosen a rigorous, but largely informal, presentation; the main contribution of the paper can be stated as follows:

Given a hierarchy, a SqPO rule, an expansive (resp. restrictive) instance of that rule into some object $O$ and factorizations for \emph{every} other object on a path from (resp. to) $O$ satisfying composability, we can \emph{uniquely} rewrite the entire hierarchy in a way that guarantees the validity of the result.

The requirement to specify all these factorizations---and also verify that they satisfy composability if necessary---can, in principle, be very onerous. Nonetheless, our experience suggests that most updates need propagate only along single edges or, at most, paths of length $2$ so that, in practice, the requirement is not too onerous. 

The other principal open question concerns the characterization of the data structures necessary to maintain an \emph{audit trail} of all updates made to a system. This would enable us to determine whether an update can be undone or not, a question that is greatly complicated by the fact of propagation, and, more generally, provide support for maintaining different \emph{versions} of the contents of a KR. This requires a major generalization of the theory of \emph{causality} between SqPO rules; see \cite{harmer2017hdr} for example. We intend to investigate this question first in the two concrete use cases discussed in this paper before attempting a full-blown generalization to arbitrary hierarchies.

\section*{References}
\bibliographystyle{splncs04}
\bibliography{ms}
\end{document}